\pdfoutput=1
\documentclass[runningheads]{llncs}
\usepackage[T1]{fontenc}
\usepackage{graphicx}
\usepackage{booktabs}
\usepackage{multirow}

\usepackage{amsfonts,amssymb}
\usepackage{amsmath}

\usepackage{geometry}
\begin{document}
\title{Adaptive Segmentation Network for Scene Text Detection}
%
%\titlerunning{Abbreviated paper title}
% If the paper title is too long for the running head, you can set
% an abbreviated paper title here
%
\author{Guiqin Zhao\inst{1*}}
\authorrunning{G. Zhao et al.}
% First names are abbreviated in the running head.
% If there are more than two authors, 'et al.' is used.
%
\institute{School of Computer Science and Technology, University of Chinese Academy of Sciences, Beijing 100049, China\\ \email{zhaoguiqin20@mails.ucas.edu.cn}\\
}

\maketitle              % typeset the header of the contribution
\begin{abstract}
Inspired by deep convolution segmentation algorithms, scene text detectors break the performance ceiling of datasets steadily. However, these methods often encounter threshold selection bottlenecks and have poor performance on text instances with extreme aspect ratios. In this paper, we propose to automatically learn the discriminate segmentation threshold, which distinguishes text pixels from background pixels for segmentation-based scene text detectors and then further reduces the time-consuming manual parameter adjustment. Besides, we design a Global-information Enhanced Feature Pyramid Network (GE-FPN) for capturing text instances with macro size and extreme aspect ratios. Following the GE-FPN, we introduce a cascade optimization structure to further refine the text instances. Finally, together with the proposed threshold learning strategy and text detection structure, we design an Adaptive Segmentation Network (ASNet) for scene text detection. Extensive experiments are carried out to demonstrate that the proposed ASNet can achieve the state-of-the-art performance on four text detection benchmarks, $i.e.$, ICDAR 2015, MSRA-TD500, ICDAR 2017 MLT and CTW1500. The ablation experiments also verify the effectiveness of our contributions.

\end{abstract}

\section{Introduction}
\label{sec:intro}
Scene text detection\cite{zhao2022explore,zhao2023flowtext,zhao2023generative} has become very popular in the computer vision community, the performance of scene text detectors on various text detection datasets is constantly refreshed. But there are still many problems to be solved to realize accurate text detection in the real-world scenarios~\cite{DBLP:journals/ijcv/LongHY21}.
Here, we focus on two general problems: discriminate segmentation threshold selection and extreme aspect ratio. 

\textbf{Discriminate segmentation threshold selection}. For the segmentation-based text detectors, the segmentation threshold used to distinguish text pixels from background pixels is indispensable. To achieve the best detection performance, researchers need to manually set different thresholds for the algorithm on different datasets. It is often very time-consuming and laborious especially when the dataset is huge, and difficult to select the appropriate threshold for open application scenarios. Therefore, it is urgently needed to obtain an appropriate segmentation threshold without performance feedback from the test set. DBNet~\cite{liao2020real} first raises the concept of adaptive threshold, but it still follows the previous method of manually selecting the threshold in the inference stage. Our method takes a further step to learn and adopt the adaptive segmentation threshold in the inference stage and achieves good performance.

\textbf{Extreme aspect ratio}. Most text detection algorithms extract text features by various convolutions. It is usually limited by the size of the receptive field, so it is difficult for them to obtain sufficient information for the accurate detection of long text instances. Most regression-based methods, such as  EAST~\cite{zhou2017east} and RRD~\cite{liao2018rotation}, face the problem of inaccurate regression caused by the lack of receptive field. Mask TextSpotter~\cite{lyu2018mask} and DeepText~\cite{zhong2017deeptext} follow a Mask R-CNN~\cite{He_2017_ICCV}-style framework to alleviate the problem of inaccurate single-stage regression due to insufficient receptive field. LOMO~\cite{zhang2019look} and MOST~\cite{he2021most} construct a cascade structure to get accurate text detection by iterative regression.  

To solve these two problems, we propose an Adaptive Segmentation Network (ASNet) for accurately detecting text in real-world scenarios. Specifically, for the first problem, we propose the discriminate segmentation threshold which contains two concepts of threshold: the dataset level threshold (DTH) and the image level threshold (ITH). The network we designed can automatically learn the dataset level threshold and learn the way to predict the image level threshold in the training process, which can be directly used in the inference stage. The performance of the threshold we learned is comparable to the performance of the manually selected threshold. For the second problem, we introduce a cascade structure to the proposed ASNet so that the network can overcome the problem of the insufficient receptive field by iteratively refining the location of the text instance. Meanwhile, we introduce the self attention mechanism to the Feature Pyramid Network (FPN)\cite{lin2017feature} to form the Global-information Enhanced Feature Pyramid Network (GE-FPN), which increases the receptive field and further improves the performance of the proposed network.
The contributions of this paper are three-fold:
\begin{itemize}
    \item We propose an Adaptive Segmentation Network (ASNet) for scene text detection that can alleviate the dilemma of segmentation threshold selection and alleviate the problem of insufficient receptive field in the text detection task.
    \item We propose the adaptive segmentation threshold which contains two concepts of learnable thresholds, namely dataset level threshold and image level threshold, which promotes the performance of the proposed network and avoids manually selecting segmentation thresholds in the inference stage.
    \item We introduce the self attention mechanism and the cascade structure to the proposed ASNet, which expands the receptive field of ASNet and further improves the performance.
\end{itemize}

\section{Methodology}
\label{sec:pagestyle}

\begin{figure}[tbp]
    \centering
    \includegraphics[width=12cm]{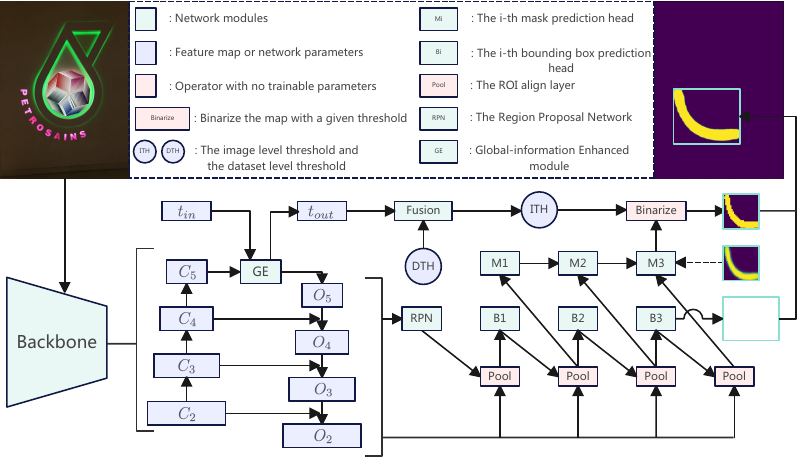}
    \caption{Workflow of the proposed ASNet in the inference stage. $C_i,i\in\{2,3,4,5\}$ denotes the multi-scale feature maps extracted by the backbone. $O_i,i\in\{2,3,4,5\}$ denotes the multi-scale feature maps output from GE-FPN. The generation of the DTH is described in Equation~\ref{eq:dth} and the fusion block and the generation of the ITH is described in Equation~\ref{eq:fusion}. }
    \label{fig:arc}
\end{figure}

\subsection{Overview}
The proposed ASNet is composed of a ResNet
50\cite{DBLP:conf/cvpr/HeZRS16} backbone, a Global-information Enhanced Feature Pyramid Network (GE-FPN), a Region Proposal Network (RPN)~\cite{NIPS2015_14bfa6bb}, and a cascade head.

The overall architecture of the proposed ASNet is shown in Fig.\ref{fig:arc}. First, we extract the multi-scale feature maps with a ResNet-50 backbone, and then the multi-scale feature maps are feed into the GE-FPN to be better aggregated. Meanwhile, the dataset level threshold (DTH) and the image level threshold (ITH) are generated in the forward propagation of the Global-information Enhanced module (GE) in the GE-FPN. Second, we input the multi-scale feature maps into the Region Proposal Network (RPN) to obtain a series of proposals, which are then input into the cascade head together with the multi-scale feature maps to obtain the score, bounding box and mask of the text instances. Then, we binarize the mask of the text instances that achieve high score with the image level threshold. Finally, We use the bounding box predicted by the last bounding box prediction head to map the binary mask back to the original size of the image and get the final detection result.

\subsection{Global-information Enhanced Feature Pyramid Network}
\begin{figure}[htbp]
    \centering
    \includegraphics[width=9cm]{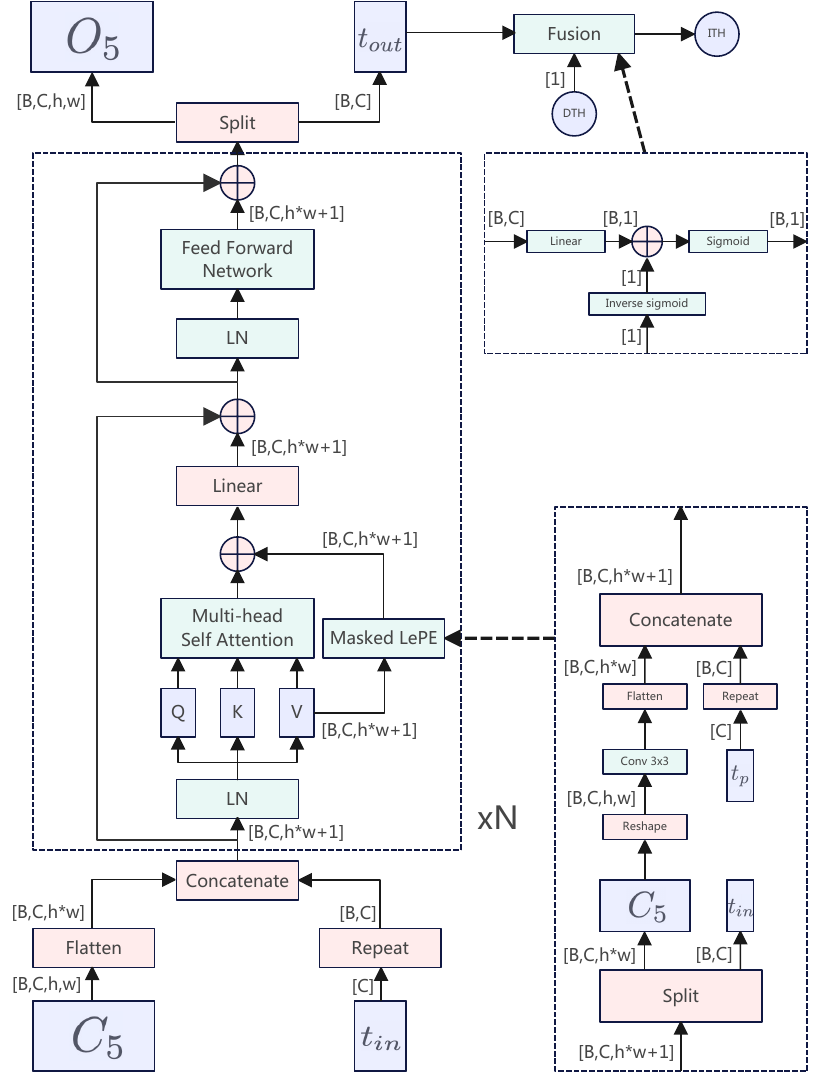}
    \caption{Architecture of the Global-information Enhanced module. For better expression, we annotate the shape of the feature map. $C_5\in \mathbb{R}^{B\times C\times h\times w},h=h_5=\frac{H}{32},w=w_5=\frac{W}{32}$, where $H,W$ are the height and width of the input image. LN denotes the Layer Normalization. }
    \label{fig:gefpn}
\end{figure}

As shown in Fig.~\ref{fig:arc}, the backbone of the proposed ASNet outputs four feature maps, we name them as $\{C_i\}_{i\in \{2,3,4,5\}}$, the size of $C_i$ is $\frac{1}{2^{i}}$ of the input image, $i.e.$, $C_i \in \mathbb{R}^{B\times C\times h_i\times w_i},h_i=\frac{H}{2^i},w_i=\frac{W}{2^i}$, where $H,W$ are the height and width of the input image. We add a Global-information Enhanced (GE) module after the lateral connection of $C_5$ to extract the global information of the image. The architecture of the GE module is shown in Fig. \ref{fig:gefpn}.  

As we can see in Fig.~\ref{fig:gefpn}, firstly, we flatten $C_5$ and combine it with $t_{in}$ by concatenation operation, where $t_{in}\in \mathbb{R}^C$ is a trainable vector of the network. Secondly, we apply Layer Normalization (LN) for the features and then feed it into Multi-head Self Attention module. Since $t_{in}$ does not have positional relationship with the pixels on $C_5$, we use Masked LePE for position encoding. Then, we apply Layer Normalization (LN) for the features and then feed it into the Feed Forward Network~\cite{vaswani2017attention}. We repeat the GE module for N times, and N is 2 in our experiments. Finally, the features encoded by GE module are split into $O_5$ and $t_{out}$, where $O_5$ is used to enhance the global information of the original FPN and $t_{out}$ is used to predict the image level threshold.

\subsection{Cascade Structure}
Most text detection algorithms based on Mask R
CNN\cite{He_2017_ICCV}, such as Mask Textspotter\cite{lyu2018mask} and DeepText\cite{zhong2017deeptext}, are faced with the problem that the single-step bounding box regression for text instance is not accurate enough. Inspired by Cascade R-CNN\cite{cai2018cascade} and HTC\cite{chen2019hybrid}, we establish a multi-step bounding box regression head. As shown in Fig.~\ref{fig:arc}, firstly, we input the proposals output by RPN into ROI align layer and obtain the text features. Secondly, we input the text features into the first bounding box head $\text{B1}$ to obtain the refined bounding box prediction. Then, we input the refined bounding box prediction into ROI align layer and obtain the text features for the second bounding box head $\text{B2}$ and the first mask head $\text{M1}$. We repeat this procedure for three times to obtain more accurate detection results. 

\subsection{Adaptive Segmentation Threshold Learning}
Segmentation threshold can be used to binarize the probability map of the text instances into the binary mask. The optimal segmentation thresholds for different images or different datasets are often different. Besides, it is very cumbersome and impractical to manually select segmentation thresholds to obtain the optimal. In this regard, we propose the adaptive discriminate threshold which contains the dataset level threshold and the image level threshold.

The dataset level threshold (DTH) is a trainable parameter of the proposed ASNet. In the training stage, the DTH for each datasets can be learned through the back propagation algorithm. In the inference stage, we fix the DTH learned in the training stage and use it as a candidate segmentation threshold for the binarization of the predicted probability map.

The image level threshold (ITH) is a value obtained by adding an offset to the DTH. Inspired by ViT\cite{dosovitskiy2020vit} and DETR\cite{carion2020end}, we obtain the predicted offset for the test image through a predefined token $t_{in}$, which is a trainable parameter of the proposed ASNet. To better explain how DTH and ITH are generated, we use $t_D$ and $t_I$ to represent the DTH and the ITH respectively, $t_{out}\in \mathbb{R}^{B\times C}$ represents the feature vector fused with global information in Fig.~\ref{fig:gefpn} and $t_d\in \mathbb{R}$ is a trainable parameter of the proposed ASNet. The generation process for $t_D$ and $t_I$ can be formulated as follows:
\begin{align}
    t_{D}&=\text{Sigmoid}(t_d)\in \mathbb{R}[0,1], \label{eq:dth}\\
    t_{I}&=\text{Sigmoid}(t_d+\text{Linear}_{C\rightarrow 1}(t_{out}))\in \mathbb{R}[0,1]^B, \label{eq:fusion}
\end{align}
where $\text{Linear}_{C\rightarrow 1}$ denotes a linear projection that project the channels of $t_{out}$ (C) to 1. 

\subsection{Adaptive Threshold Loss}

We propose a novel Adaptive Threshold Loss to supervise both the text instance probability maps and the two-level segmentation thresholds at the same time. The loss function of mask branch for a single text instance which has matched the ground truth of a certain text instance can be formulated as follows:\\
\begin{align}
    \mathcal{L}_{\text{ath}}(m,\hat{m},t_D,t_I)&=\mathcal{L}_{\text{SCEL}}(m,\hat{m})\nonumber\\
    &+\alpha \mathcal{L}_{\text{DL}}(\text{Step}(m,t_D),\hat{m})+\beta \mathcal{L}_{\text{DL}}(\text{Step}(m,t_I),\hat{m}),\\
    \mathcal{L}_{\text{SCEL}}(m,\hat{m}) &= -\frac{1}{h_mw_m} \sum_{i=1}^{h_m}\sum_{j=1}^{w_m}[\hat{m}_{ij}\text{log} m_{ij}+(1-\hat{m}_{ij})\text{log}(1-m_{ij})],\\
    \mathcal{L}_{\text{DL}}(x,y)&= 1-\frac{2\cdot\text{sum}(xy)}{\text{sum}(x)+\text{sum}(y)},\\
    % m_{D}&=\text{Step}(m,t_D),\\
    % m_{I}&=\text{Step}(m,t_I),\\
    \text{Step}(x,y)&=\frac{1}{1+\text{exp}[-k(x-y)]},\label{eq:8}
    % &m_{bd}=\frac{1}{1+exp(-k*(m_p-t_1))}\\
    % &m_{bi}=\frac{1}{1+exp(-k*(m_p-t_2))}\\
\end{align}
where $m\in \mathbb{R}[0,1]^{h_m\times w_m}$ is the predicted probability map of the mask branch, $\hat{m}\in \mathbb{Z}[0,1]^{h_m\times w_m}$ is the ground truth of a certain text instance mask which matches this proposal. $\alpha$ and $\beta$ are used to balance the importance of the three losses, we set $\alpha=\beta=0.5$ in our experiments. The $\text{Step}$ function equals to the $\text{Step}$ function in DBNet\cite{liao2020real}, we set $k=50$ in our experiments. As proved in DBNet, the $\text{Step}$ function is a continuous approximation of the $\text{Binarize}$ function, $i.e.$, $\text{Binarize}(m,t) = \text{I}[m>t]$, where $\text{I}$ is the characteristic function. In the training stage, we use $\text{Step}$ function to replace $\text{Binarize}$ function so that the network can obtain the ability of predicting the image level segmentation threshold for each image through backward propagation.

\subsection{Optimization}
The loss function for the proposed ASNet can be formulated as:
\begin{equation} \mathcal{L}=\mathcal{L}_{\text{RPN}}+\sum_{i=1,2,3}\lambda_i(\mathcal{L}_{\text{score},i}+\gamma_b \mathcal{L}_{\text{bbox},i}+\gamma_m \mathcal{L}_{\text{mask},i}),  
\end{equation} 
where $\mathcal{L}_{\text{RPN}}$ denotes the loss for the Region Proposal Network, $\mathcal{L}_{\text{score},i}$ and $\mathcal{L}_{\text{bbox},i}$ denote the text score loss and bounding box regression loss for the $i$-th bounding box prediction head $\text{B}i$ in Fig~\ref{fig:arc}, $i = 1,2,3$.  $\mathcal{L}_{\text{mask},i}$ is the mask loss for the $i$-th mask prediction head $\text{M}i$ in Fig~\ref{fig:arc}.

\section{Experiments}
We conducted quantitative experiments on four public benchmarks: ICDAR 2015\cite{karatzas2015icdar}, MSRA-TD500~\cite{yao2012detecting}, ICDAR 2017\cite{nayef2017icdar2017} and CTW1500~\cite{DBLP:journals/pr/LiuJZLZ19} to verify the effectiveness of the proposed algorithm and the effectiveness of the proposed components.

\subsection{Implementation Details}
The backbone of the proposed method is the ResNet-50 which is pre-trained on ImageNet\cite{DBLP:conf/cvpr/DengDSLL009}. For experiments that introduce additional data, the proposed ASNet is first pre-trained on Cocotext v2\cite{veit2016cocotext} for 80 epochs, then it is fine-tuned separately on the dataset which we report the results for 160 epochs. For experiments that do not introduce additional data, we train ASNet for 160 epochs on ICDAR 2015 and CTW1500 respectively. We train 1200 epochs on MSRA-TD500.  The data augmentation contains color jitter, random rotation, random horizontal flipping, random scale, random crop, and padding to 800×800. In the inference stage, we resize the long dimension of test images to 640, 1920, 1600 and 640 for MSRA-TD500, ICDAR2015, ICDAR2017 and CTW1500 respectively.  

\subsection{Straight Text Detection Results}
To verify the detection performance of the proposed ASNet on straight text, we make an extensive comparison with the recent state-of-the-art methods on three text detection datasets, $i.e.$, ICDAR 2015, MSRA-TD500 and ICDAR 2017.  

\begin{table*}[!htbp]
\setlength{\tabcolsep}{7pt}
\centering
\caption{Comparison with recent state-of-the-art methods on ICDAR 2015\cite{karatzas2015icdar} and MSRA-TD500\cite{yao2012detecting}. We fixed the backbone of all the compared methods to ResNet-50. 'Ext.' denotes extra training data, $\dagger$ denotes using DCN\cite{dai2017deformable} in backbone. }

\begin{tabular}{ccc|ccc|ccc}
\toprule
\multirow{2}{*}{Methods} & \multirow{2}{*}{Paper} & \multirow{2}{*}{Ext.} & \multicolumn{3}{c|}{ICDAR2015} & \multicolumn{3}{c}{MSRA-TD500}\tabularnewline
% \cline{4-9} 
 &  &  & R(\%) & P(\%) & F(\%) & R(\%) & P(\%) & F(\%) \tabularnewline
\midrule
PAN\cite{wang2019efficient} & ICCV'19 &  & 77.8 & 82.9 & 80.3 & 77.3 & 80.7 & 78.9\tabularnewline

PAN\cite{wang2019efficient} & ICCV'19 & \checkmark & 81.9 & 84.0 & 82.9 & 83.2 & 85.7 & 84.5\tabularnewline

PSENet\cite{wang2019shape} & CVPR'19 & \checkmark & 84.5 & 86.9 & 85.7 & - & - & -\tabularnewline

CRAFT\cite{baek2019character} & CVPR'19 & \checkmark & 84.3 & 89.8 & 86.9 & 78.2 & 88.2 & 82.9\tabularnewline

LOMO\cite{zhang2019look} & CVPR'19 & \checkmark & 83.5 & 91.3 & 87.2 & - & - & -\tabularnewline

DB$\dagger$\cite{liao2020real} & AAAI'20 & \checkmark & 83.2 & 91.8 & 87.3 & 79.2 & 91.5 & 84.9\tabularnewline

% ContourNet\cite{wang2020contournet} & CVPR'20 & \Checkmark{} & 86.1 & 87.6 & 86.9 & - & - & - & - & - & -\tabularnewline
% \hline 
DRRG\cite{zhang2020deep} & CVPR'20 & \checkmark & 84.7 & 88.5 & 86.6 & 82.3 & 88.1 & 85.1\tabularnewline

FCENet\cite{zhu2021fourier} & CVPR'21 &  & 84.2 & 85.1 & 84.6 & - & - & -\tabularnewline

MOST\cite{he2021most} & CVPR'21 & \checkmark & 87.3 & 89.1 & 88.2 & 82.7 & 90.4 & 86.4\tabularnewline

TextBPN\cite{DBLP:conf/iccv/Zhang0YWY21} & ICCV'21 & \checkmark & - & - & - & 84.5 & 86.6 & 85.6\tabularnewline
\midrule
ASNet & Ours &  & $83.5$ & $90.0$ & $86.6$ & 76.3 & $\mathbf{95.3}$ & 84.7\tabularnewline
ASNet & Ours & \checkmark & $\mathbf{88.9}$ & $\mathbf{91.9}$ & $\mathbf{90.4}$ & $\mathbf{88.1}$ & 93.1 & $\mathbf{90.6}$\tabularnewline
\bottomrule
\end{tabular}
\label{tab:ic15}
\vspace{-3mm}
\end{table*}

\subsubsection{ICDAR 2015} mainly focuses on multi-oriented text in natural scenes. As shown in Table~\ref{tab:ic15}, the proposed ASNet achieves SOTA performance on ICDAR2015. If compared with the methods without pre-training, the proposed ASNet outperforms the previous SOTA method FCENet by 2.0\% (86.6\% $vs.$ 84.6\%) in F-measure. If compared with the algorithm with pre-training, the proposed ASNet can still outperforms the previous SOTA method MOST by 2.2\% (90.4\% $vs.$ 88.2\%) in F-measure.
The result shows that the proposed ASNet can achieve good detection performance for the text instances in natural scenes.

\subsubsection{MSRA-TD500} is a multilingual dataset containing many extreme long text instances. As shown in Table~\ref{tab:ic15}, the proposed ASNet achieves SOTA performance on MSRA-TD500. If compared with the methods without pre-training, the proposed ASNet outperforms the previous SOTA method PCR by 2.3\% (84.7\% $vs.$ 82.4\%) in F-measure. If compared with the algorithm with pre-training, the proposed ASNet can still outperforms the previous SOTA method GNNets by 2.1\% (90.6\% $vs.$ 88.5\%) in F-measure. The result shows that the proposed ASNet is capable of detecting text instances with extreme aspect ratio.

\subsubsection{ICDAR 2017} is a dataset containing multilingual text instances.  As shown in Table~\ref{tab:ic15}, the proposed ASNet achieves SOTA performance on ICDAR 2017. The result shows that the proposed ASNet can detect text instances of different languages well.

\begin{table*}[htbp]
\setlength{\tabcolsep}{7pt}
\centering
\caption{Comparison with recent state-of-the-art methods on ICDAR 2017\cite{nayef2017icdar2017} test set. }

\begin{tabular}{ccc|ccc}
\toprule
\multirow{2}{*}{Methods} & \multirow{2}{*}{Paper} & \multirow{2}{*}{Ext.} & \multicolumn{3}{c}{ICDAR2017}\tabularnewline
% \cline{4-6} 
 &  &  & R(\%) & P(\%) & F(\%)\tabularnewline
\midrule

PSENet\cite{wang2019shape} & CVPR'19 & \checkmark & 68.2 & 73.8 & 70.9\tabularnewline

CRAFT\cite{baek2019character} & CVPR'19 & \checkmark & 68.2 & 80.6 & 73.0\tabularnewline

LOMO\cite{zhang2019look} & CVPR'19 & \checkmark & 60.6 & 78.8 & 68.5\tabularnewline
 
DB$\dagger$\cite{liao2020real} & AAAI'20 & \checkmark & 67.9 & 83.1 & 74.7\tabularnewline

DRRG\cite{zhang2020deep} & CVPR'20 & \checkmark & 61.0 & 75.0 & 67.3 \tabularnewline

MOST\cite{he2021most} & CVPR'21 & \checkmark & 72.0 & 82.0 & 76.7\tabularnewline
\midrule
ASNet & Ours & \checkmark & $\mathbf{74.5}$ & $\mathbf{84.2}$ & $\mathbf{79.1}$\tabularnewline
\bottomrule
\end{tabular}
\label{tab:ic17}
% \vspace{-3mm}
\end{table*}

\subsection{Curve Text Detection Results}
CTW1500 is a widely used dataset for arbitrarily-shaped text detection. As shown in Table~\ref{tab:ctw}, the proposed ASNet achieves SOTA performance on CTW1500. If compared with the methods without pre-training, the proposed ASNet outperforms the previous SOTA method TextBPN by 0.3\% (84.3\% $vs.$ 84.0\%) in F-measure. If compared with the methods with pre-training, the proposed ASNet outperforms the previous SOTA method TextBPN by 0.2\% (85.2\% $vs.$ 85.0\%) in F-measure. 
% Some detection results of the proposed ASNet on CTW1500 are shown in Fig~\ref{fig:ctw}. 

\begin{table*}[!htbp]
\setlength{\tabcolsep}{7pt}
\centering
\caption{Comparison with recent state-of-the-art methods on CTW1500. }

\begin{tabular}{ccc|ccc}
\toprule
\multirow{2}{*}{Methods} & \multirow{2}{*}{Paper} & \multirow{2}{*}{Ext.} & \multicolumn{3}{c}{ICDAR2017}\tabularnewline
% \cline{4-6} 
 &  &  & R(\%) & P(\%) & F(\%)\tabularnewline
\midrule
DRRG\cite{zhang2020deep} & CVPR'20& \checkmark & 83.0 & 85.9 & 84.5\tabularnewline

FCENet\cite{zhu2021fourier} & CVPR'21 & & 80.7 & 85.7 & 83.1\tabularnewline

TextBPN\cite{DBLP:conf/iccv/Zhang0YWY21} & ICCV'21 & & 80.6 & 87.7 & 84.0\tabularnewline

TextBPN\cite{DBLP:conf/iccv/Zhang0YWY21} & ICCV'21 & \checkmark & 83.6 & 86.5 & 85.0\tabularnewline

\midrule
ASNet & Ours & & $82.4$ & $86.2$ & $84.3$\tabularnewline
ASNet & Ours & \checkmark & $82.4$ & $\mathbf{88.1}$ & $\mathbf{85.2}$\tabularnewline
\bottomrule
\end{tabular}
    \label{tab:ctw}
\vspace{-3mm}
\end{table*}

% \begin{figure}[!htbp]
%     \centering
%     \caption{Some detection results of the proposed ASNet on CTW1500 test set.}
%     \includegraphics[width=12cm]{Figures/ctw.pdf}
%     \label{fig:ctw}
% \end{figure}
\subsection{Ablation study}
We conducted ablation studies on the ICDAR 2015 dataset, shown in Table \ref{tab:main}. The baseline of the proposed ASNet is Mask R-CNN, whose result is referred from open-mmocr\cite{mmocr2021}.  

\begin{table}[!htbp]
\setlength{\tabcolsep}{7pt}
\centering
\caption{Ablation studies on ICDAR 2015 under the protocol of IoU@0.5. The Aug in the table denotes data augmentations we used in the training stage.}
\begin{tabular}{ccccc|ccc}
\toprule
Cascade & DTH & ITH & GE & Aug & R(\%) & P(\%) & F(\%)\tabularnewline
\midrule
 &  &  &  &  & 78.30 & 87.20 & 82.50\tabularnewline

\checkmark &  &  &  &  & 78.43 & \textbf{90.75} & 84.14\tabularnewline

 & \checkmark &  &  &  & 82.67 & 86.63 & 83.41\tabularnewline

 & \checkmark & \checkmark &  &  & 80.84 & 87.68 & 84.12\tabularnewline

 &  &  & \checkmark &  & 81.46 & 86.46 & 83.89\tabularnewline

 & \checkmark & \checkmark & \checkmark &  & 81.42 & 88.07 & 84.61\tabularnewline

\checkmark & \checkmark & \checkmark & \checkmark &  & 80.98 & 90.58 & 85.51\tabularnewline

\checkmark & \checkmark & \checkmark & \checkmark & \checkmark & \textbf{83.53} & 89.99 & \textbf{86.64}\tabularnewline
\bottomrule
\end{tabular}

\label{tab:main}
\end{table}

\subsubsection{Cascade structure} As shown in Table \ref{tab:main}, compared with the baseline, the cascade structure can bring relative improvements of 3.55\% (90.75\% $vs.$ 87.20\%) in Precision on ICDAR 2015 while keep the Recall nearly unchanged (78.43\% $vs.$ 78.30\%), which shows that the multi-step bounding box regression process can significantly improves the precision of detecting text instances.  

\subsubsection{Global-information enhanced module} As shown in Table~\ref{tab:main}, the addition of the Global-information Enhanced module after the lateral connection of $C_5$ of the FPN brings relative improvements of 3.16\% (81.46\% $vs.$ 78.30\%) in Recall and 1.39\% (83.89\% $vs.$ 82.50\%) in F-measure on ICDAR 2015. The results shows that GE module can effectively enhance the receptive field of the original FPN and makes the network capable of detecting text instances with extreme aspect ratio.

\subsubsection{Adaptive segmentation threshold}
\begin{figure}[tbp]
    \centering
    \caption{Performance comparison between the image level threshold (ITH) and manually selected threshold (MTH) on the training set and test set of ICDAR 2015. TH denotes the segmentation threshold used to binarize the predicted text instance probability map.}
    \includegraphics[width=12cm]{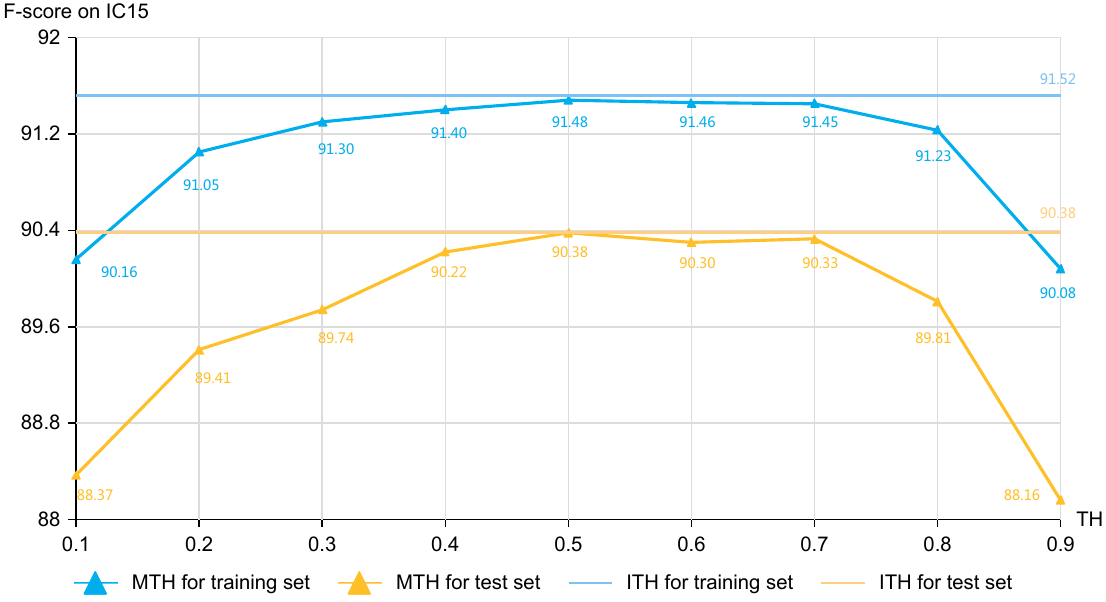}
    \label{fig:dth}
\end{figure}

As shown in Table~\ref{tab:main}, with DTH, the model achieve relative improvements of 0.91\% (83.41\% $vs.$ 82.50) in F-measure on ICDAR 2015. With ITH, the model can bring relative improvements of 0.71\% (84.12\% $vs.$ 83.41\%), which shows that adaptive threshold is more effective than a fixed threshold. Moreover, in order to verify the effectiveness of the adaptive segmentation threshold in the inference stage, we compared the F-measure performance of the model using the image level threshold and the manually selected threshold at an interval of 0.1 on ICDAR 2015. As shown in Fig~\ref{fig:dth}, When testing on the training set or on the testing set, ITH can achieve better or equivalent performance. 

\section{Conclusions}
\label{sec:majhead}
In this paper, we propose to automatically learn the adaptive segmentation threshold for segmentation-based scene text detectors, which can not only assist the training process of the network but also help to select segmentation threshold in the inference stage. Meanwhile, we introduce the self attention mechanism into our method and propose the Global-information Enhanced Feature Pyramid Network (GE-FPN), which expands the receptive field of the network and improves the performance of the network. The experiments demonstrate that the proposed ASNet achieves state-of-the-art performance on ICDAR 2015, MSRA-TD500, ICDAR 2017 MLT and CTW1500.

% ---- Bibliography ----
%
% BibTeX users should specify bibliography style 'splncs04'.
% References will then be sorted and formatted in the correct style.
%
\bibliographystyle{splncs04}
\nocite{*}
\bibliography{reference}
 
\end{document}